\documentclass[]{spie}  %>>> use for US letter paper
\usepackage{amsmath,amsfonts,amssymb}
\usepackage{graphicx}
\usepackage[colorlinks=true, allcolors=blue]{hyperref}

\title{Deep Transductive Transfer Learning for Automatic Target Recognition}

\author[a]{Shoaib M. Sami}
\author[a]{Nasser M. Nasrabadi}
\author[b]{Raghuveer Rao}
\affil[a]{LCSEE Dept., West Virginia University, Morgantown, WV, USA}
\affil[b]{DEVCOM Army Research Laboratory, Adelphi, MD, USA}

\authorinfo{Further author information: (Send correspondence to S. M. S.)\\S. M. S.: E-mail: sms00052@mix.wvu.edu, Telephone: +1 304 435 7049\\N. M. N. : E-mail: nasser.nasrabadi@mail.wvu.edu, Telephone: +1 304 293 4815\\ This paper is accepted in: SPIE Defense \& Commercial Sensing 2023, Conference 12521, Automatic target recognition XXXIII, Orlando, Florida}

\begin{document} 
\maketitle

\begin{abstract}
One of the major obstacles in designing an automatic target recognition (ATR) algorithm, is that there are often labeled images in one domain (i.e., infrared source domain) but no annotated images in the other target domains (i.e., visible, SAR, LIDAR). Therefore, automatically annotating these images is essential to build a robust classifier in the target domain based on the labeled images of the source domain. Transductive transfer learning is an effective way to adapt a network to a new target domain by utilizing a pretrained ATR network in the source domain. We propose an unpaired transductive transfer learning framework where a CycleGAN model and a well-trained ATR classifier in the source domain are used to construct an ATR classifier in the target domain without having any labeled data in the target domain. We employ a CycleGAN model to transfer the mid-wave infrared (MWIR) images to visible (VIS) domain images (or visible to MWIR domain). 
To train the transductive CycleGAN, we optimize a cost function consisting of the adversarial, identity, cycle-consistency, and categorical cross-entropy loss for both the source and target classifiers. In this
paper, we perform a detailed experimental analysis on the challenging DSIAC ATR dataset. The dataset consists of ten classes of vehicles at different poses and distances ranging from 1-5 kilometers on both the MWIR and VIS domains. In our experiment, we assume that the images in the VIS domain are the unlabeled target dataset. We first detect and crop the vehicles from the raw images and then project them into a common distance of 2 kilometers. Our proposed transductive CycleGAN achieves 71.56\%  accuracy in classifying the visible domain vehicles in the DSIAC ATR dataset. 
\end{abstract}

% Include a list of keywords after the abstract 
\keywords{Automatic Target Recognition, Transductive Transfer Learning, CycleGAN, Domain Adaptation.}

\section{INTRODUCTION}
\label{sec:intro}  % \label{} allows reference to this section
Intelligent automatic target recognition (ATR) algorithms are necessary for advanced military applications where they can drastically reduce civilian casualties. ATR algorithms are expected to automatically detect and classify military vehicles\cite{nasser_deeptarget} with very little human supervision. To build a robust ATR algorithm, a large-scale target vehicle dataset is required. Generally, ATR images are captured in the visible, mid-wave infrared (MWIR), long-wave infrared (LWIR), or  radar (i.e., SAR) domain.  There are real-world scenarios where there is well-labeled military vehicle data in one domain (source domain), where an ATR classifier can easily be designed,  but no labeled data of the same vehicles are available in another domain (target domain) where an ATR classifier is urgently needed. Also, annotating unlabeled data is costly, time-consuming, and a complicated task. Therefore, constructing an ATR classifier in the unlabeled target domain without labeling the data is a challenging and essential task that can reduce the cost and complexity of the ATR vehicle annotation and classifier design process.\cite{atr1,atr2_review,atr3,atr4}

A large number of algorithms have been proposed in the literature to annotate unlabeled data. They can be broadly classified into two categories: self-supervised and semi-supervised \cite{simclr_self_supervised}. Transductive transfer learning \cite{transductive_eccv_ref,transductive_nips_ref} is one type of semi-supervised method to annotate unlabelled data in the target domain using labeled information from the source domain. In the literature, transductive transfer learning is widely used in different applications \cite{Utkin2018-gg, Luo2022-ld}. However, to the best of our knowledge, transductive transfer learning-based ATR vehicle annotation has not been exploited. Therefore, there is a scope to annotate the ATR vehicles using a transductive transfer learning approach. 

 In this paper, we propose a CycleGAN-based transductive transfer learning approach with a CNN-based ATR classifier in the source and target domains. CycleGAN \cite{cyclegan}, DRIT \cite{drit}, and MUNIT \cite{munit} etc. are popular unpaired image-to-image translation algorithms \cite{review_i2i}. In this work, we use the CycleGAN network because it is effective and widely used for domain translation tasks. The goal of the proposed transductive transfer learning network is to construct a target domain classifier with no labeled data. We analyze the performance of both the source and the target domain classifiers on the DSIAC ATR dataset \cite{dsiac}.
The organization of this paper is as follows: we discuss the literature review of our related work in Section 2. We also describe the methodology and experiment details in Sections 3 and 4, respectively. The result, discussion, and conclusion are elicited in Sections 5 and 6, respectively.
 
\section{Related Works}
\subsection{Automatic Target Recognition}

The automatic target recognition process can be divided into two major tasks: (a) target vehicle detection and (b) target classification. Deep convolutional neural networks such as Faster R-CNN \cite{rcnn}, SSD \cite{ssd}, YOLO \cite{yolo}, etc.,  have been used to detect different military vehicles in several benchmark ATR datasets.\cite{nasser_deeptarget}.  Target classification algorithms can also be divided into two major categories: i) feature-based and ii) model-based. The model-based ATR algorithms perform better as compared to feature-based methods. The model-based deep learning ATR classification algorithms are generally based on the convolutional neural networks (CNNs) \cite{nasser_deeptarget}, generative adversarial networks (GANs) \cite{semi-supervised_gan}, recurrent neural networks (RNN), long short-term memory (LSTM) \cite{rnn_lstm}, or autoencoders \cite{autoencoder_deng}. 
Nasrabadi et al. \cite{nasser_deeptarget} proposed a CNN-based network to classify forward-looking infrared images in the LWIR Comanche dataset \cite{comanche}. In \cite{meta_domenic} authors proposed Meta-UDA for detecting unlabelled mid-wave infrared targets by using the labeled visible domain images in the DSIAC ATR dataset \cite{dsiac}.  Ding et al. \cite{noise_sar_ding} used different types of augmentation including translation, speckle noise augmentation, and generating pose synthetic images to classify SAR images using a CNN classifier. Furthermore, to classify the noisy SAR images, Wang et al. \cite{decouple_neural_noise} used a
coupled network aggregating a despeckling subnetwork and a CNN classifier.

 The classical CNN classifiers need a large
number of training samples; to address this problem, few-shot learning is employed for the ATR vehicle classification. Bi-LSTM-based prototypical few-shot learning \cite{few_shot_bilstm}, hybrid inference \cite{hybrid_inference_wang2021}, meta-learning \cite{few_shot_meta} based few-shot
learning is used to classify SAR images in the MSTAR dataset. Also, autoencoder and supervised constraint are
used to classify the small samples in the ATR dataset \cite{autoencoder_deng}. Moreover, a semi-supervised ATR recognition method is
proposed to classify the SAR images \cite{semi-supervised_gan}. The authors \cite{semi-supervised_gan} constructed a CNN-based ATR classifier by using original
target images as well as synthetic images generated by GAN.

\subsection{Transductive Transfer Learning}

Transductive transfer learning (TTL) is widely used in the literature for classifying unlabelled data. For example, the authors \cite{Marcacini2018-co} propose a cross-domain aspect label propagation-based transductive learning method for opinion mining and sentiment analysis. RNN and LSTM-based transductive learning method is proposed for
optical character recognition \cite{He2018-ql}. This approach considers that both the OCR and the text contain common information in the high dimensional space. The emotion recognition in the wild dataset is solved by sparse transductive transfer linear discriminant analysis \cite{Zong2016-nb}. The transductive deep transfer learning approach based on VGGFace16-Net is
proposed for cross-domain expression recognition; in this work, cross-entropy loss of the source domain and regression loss of the target domain is optimized \cite{Yan2019-gi}. TTL is used to detect zero-day attacks in intrusion detection systems. This method uses cluster correspondence in the domain adaptation using manifold alignment \cite{Sameera2020-bq}. Supervised TTL is used for gesture classification by using the electromyographic dataset \cite{Kobylarz2020-qt}. Takagi-Sugeno-Kang's fuzzy logic systems based on transductive transfer learning are used for detecting epilepsy using electroencephalogram signals \cite{Deng2018-ge, Yang2016-ww}. Genetic programming and distributional correspondence indexing-based transductive transfer learning is used for document and text classification \cite{Fu2017-bn,Moreo2021-bj}.  The crop classification is essential for agriculture and food management, but sometimes there is only a small amount of ground truth data. To solve this issue, authors used a transductive transfer learning-based approach \cite{Luo2022-ld} for crop classification. Deep Forest-based TTL
is used in cross-domain transfer learning and for accuracy measures into the MNIST, USPS, Amazon, DSLR, Webcam, and Caltech-256 datasets  \cite{Utkin2018-gg}.

\section{Methodology}
\begin{figure} [ht]
   \begin{center}
   \begin{tabular}{c} %% tabular useful for creating an array of images 
   \includegraphics[scale = 0.78]{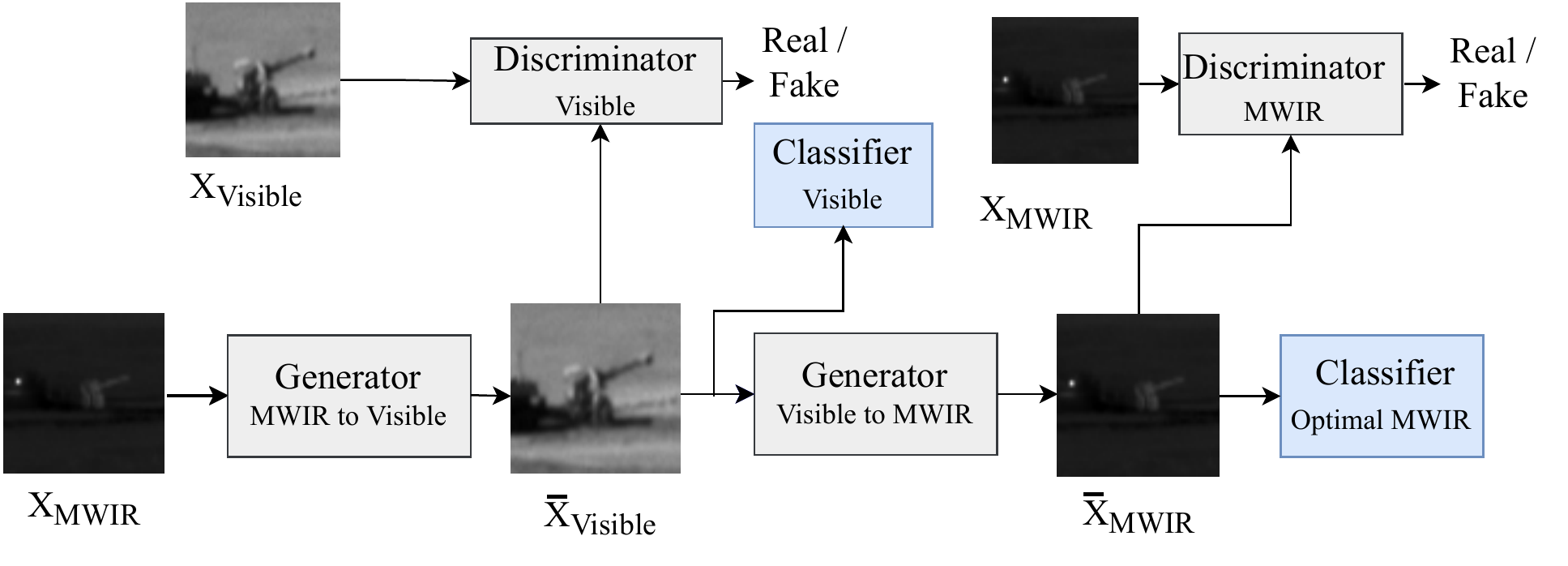}
   \end{tabular}
   \end{center}
   \caption[example] 
%>>>> use \label inside caption to get Fig. number with \ref{}
   { \label{fig:example1} 
Block diagram of the deep transductive transfer learning framework for automatic target classification.}
   \end{figure}

In this section, we will describe our proposed transductive CycleGAN. The first generator of our CycleGAN network transforms the source domain images into the target domain images. Then, the second CycleGAN generator transforms the synthetic images back into the source domain images. Moreover, there are dedicated source and target domain CNN-based ATR classifiers that backpropagate the categorical cross-entropy loss to help to train the whole TTL network. Our proposed method is depicted in 
 \autoref{fig:example1}.

\subsection{Generative Adversarial Network (GAN)}
GAN \cite{gan} is a powerful tool for different types of image generation, image translation, and super-resolution. A generator and a discriminator are the building blocks of a GAN architecture. The GAN generates images from a noise variable $z$; the generator function can be depicted as $G : z\to y$. The generator $G(z;\theta_{g})$ is optimized for the $\theta_{g}$ parameter. On the other hand, the discriminator function $D$ is designed to distinguish between the real image distribution and the generated image distribution. In practice, the discriminator is a simple binary classifier that classifies whether the image is real (y) or fake (G(z)). In GAN, the generator and the discriminator play with each other as a two-player mini-max game\cite{gan}. The objective function of GAN can be expressed by $\mathcal{L}(G,D)$:
\begin{equation}\label{eq:1}
  \begin{aligned}
\mathcal{L}(G,D) = \min_{G}\max_{D}\mathbb{E}_{x\sim p_{\text{data}}(x)}[\log{D(x)}] +  \mathbb{E}_{z\sim p_{\text{z}}(z)}[1 - \log{D(G(z))}].
  \end{aligned}
\end{equation}

\subsection{CycleGAN}
In the unpaired image-to-image translation literature, CycleGAN \cite{cyclegan} has been successfully employed as a powerful domain-to-domain transformation tool. This algorithm consists of two generators and two discriminators. The first generator (G) translates the domain X images into the domain Y images. The second generator (F) translates the domain Y images into the domain X images. The combined adversarial loss for the CycleGAN's generators G and F can be expressed by $\mathcal{L}_{GAN}(G, F, D_x, D_y)$:
\begin{equation}\label{eq:2}
  \begin{aligned}
    \mathcal{L}_{GAN}(G,F,D_x,D_y) = \min_{G}\max_{D_y}\mathbb{E}_{y\sim p_{\text{data}}(y)}[\log{D_y(y)}] +  \mathbb{E}_{x\sim p_{\text{data}}(x)}[1 - \log{D_y(G(x))}]\\ +  \min_{F}\max_{D_x}\mathbb{E}_{x\sim p_{\text{data}}(x)}[\log{D_x(x)}] +  \mathbb{E}_{y\sim p_{\text{data}}(y)}[1 - \log{D_x(F(y))}].
  \end{aligned}
\end{equation}
This mapping does not always depict a purposeful way. So, the authors proposed that the mapping should be a bijection. If there is no bijection mapping then the generator (G) can map an image in an infinite way that follows a similar distribution of domain Y. In the CycleGAN paper, the authors introduced the cycle-consistency loss, which translates the domain X image to the domain Y image then the domain Y translated image is converted back to the domain X image by the second generator F. Mathematically; it can be described as a generator $ G: X \to Y$ and another generator $ F : Y \to X$. Also, the generators G and F should translate in the inverse domain, and mapping should be bijection \cite{cyclegan}. The Cycle consistency loss can be denoted as $\mathcal{L}_{Cycle}$ where $\eta_{1}$ and $\eta_{2}$ are the hyper-parameters:
\begin{equation}\label{eq:3}
  \begin{aligned}
\mathcal{L}_{Cycle}(G,F) = \eta_{1}*\mathbb{E}_{x\sim p_{\text{data}}(x)}\left( \left\|F(G(x))-x  \right\|_{1} \right)+\eta_{2}*\mathbb{E}_{y\sim p_{\text{data}}(y)}\left( \left\|G(F(y))-y  \right\|_{1} \right) .
  \end{aligned}
\end{equation}

\begin{figure} [b]
\begin{center}
\begin{tabular}{c} %% tabular useful for creating an array of images 
\includegraphics[scale = 0.65]{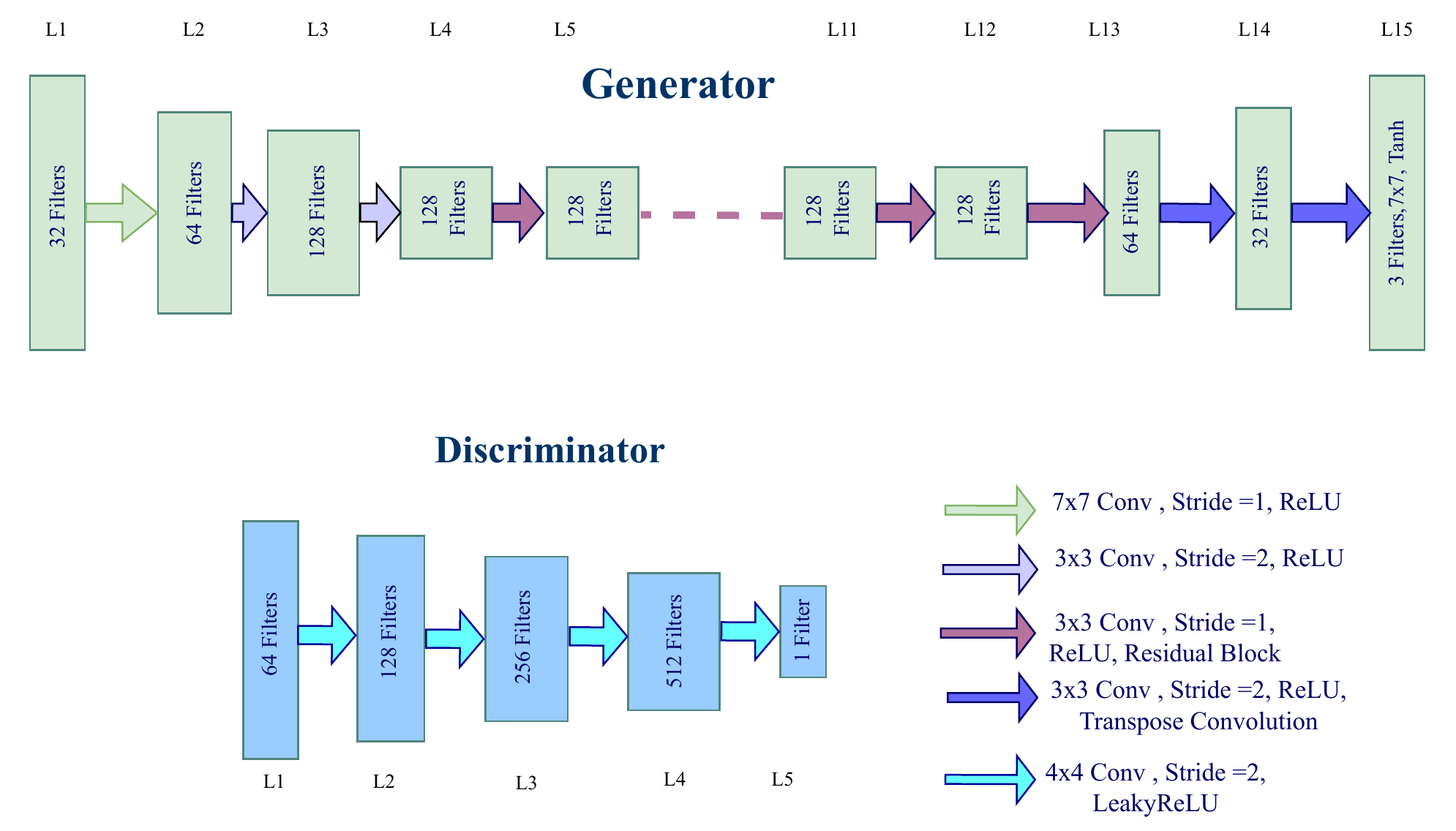}
\end{tabular}
\end{center}
\caption[example] 
%>>>> use \label inside caption to get Fig. number with \ref{}
{ \label{fig:example2} 
Block diagram of the CycleGAN's generator and discriminator.\cite{cyclegan_figure_researchgate}}
\end{figure} 

Moreover, the CycleGAN generator has two vehicle identity mapping regularizers that are employed as real images on the target domain and fed as input for the source generator. This loss is called identity loss ($\mathcal{L}_{identity}$) where $\eta_{3}$ and $\eta_{4}$ are the hyper-parameters:
\begin{equation}\label{eq:4}
  \begin{aligned}
  \mathcal{L}_{Identity}(G,F) = \eta_{3}*\mathbb{E}_{y\sim p_{\text{data}}(y)}\left( \left\|G(y)-y  \right\|_{1} \right)+\eta_{4}*\mathbb{E}_{x\sim p_{\text{data}}(x)}\left( \left\|F(x)-x  \right\|_{1} \right) .
    \end{aligned}
\end{equation}

The total CycleGAN loss ($\mathcal{L}_{CycleGAN}$) is the summation of the adversarial loss, cycle consistency loss, and identity loss where $\lambda_{a}$, $\lambda_{b}$, and $\lambda_{c}$ are the hyper-parameters: 
\begin{equation}\label{eq:5}
  \begin{aligned}
  \mathcal{L}_{CycleGAN}(G,F,D_x,D_y) =\lambda_{a}* \mathcal{L}_{GAN}(G,F,D_x,D_y) +\lambda_{b}* \mathcal{L}_{Cycle}(G,F) + \lambda_{c}*\mathcal{L}_{Identity}(G,F) .
    \end{aligned}
\end{equation}

\subsection{Source and Target Domain Classifiers}
The transductive CycleGAN network shown in \autoref{fig:example1}, has source and target domain ATR classifiers. In our experiment, the architecture of both classifiers is ResNet-18 \cite{resnet}. The skip connection in ResNet makes it possible to build a deeper network and helps to backpropagate loss. In the proposed method, we optimize the cross-entropy loss ($\mathcal{L}_{CE}$) of the classifiers. In this equation,
$y_{o,c}$ is the true probability and $p_{o,c}$ is the predicted probability of the target chip in the training data. Here, the class number is denoted by c.
\begin{equation}\label{eq:6}
  \begin{aligned}
   \mathcal{L}_{CE} = -\sum_{c=1}^{10}y_{o,c}\log(p_{o,c}) .
    \end{aligned}
\end{equation}

\subsection{Transductive CycleGAN}
Our proposed transductive CycleGAN is depicted in \autoref{fig:example1}, which consists of a CycleGAN and two residual network-based CNN classifiers. The architecture of the transductive CycleGAN generators and discriminators is illustrated in \autoref{fig:example2}. The transductive CycleGAN generator consists of fractionally-strided convolutions, regular convolutions, and residual blocks.  

In this experiment, we consider the MWIR and VIS domain ATR vehicle datasets as the source and target domain data, respectively. Moreover, in this experiment, the source domain classifier is assumed to be well-trained  (called the optimal MWIR classifier) and its weights are fixed. During the training, the source domain classifier's weights are initialized to the target domain classifier's weights which are gradually trained using the transductive transfer learning process. The proposed network's weights are obtained by minimizing the adversarial loss, cycle consistency loss, identity loss of CycleGAN, and categorical cross-entropy loss of the source and target domain ATR classifiers. It should be be pointed out that the source domain classifier loss is also backpropagated to the transductive network, but its weights are not updated during the training of the transductive network. The total loss of the transductive CycleGAN model is $\mathcal{L}_{Total}$, given by: 

\begin{equation}\label{eq:7}
  \begin{aligned}
  \mathcal{L}_{Total} = \mathcal{L}_{CycleGAN} +\lambda_{CE}* \mathcal{L}_{CE-Source} + \lambda_{CE}*\mathcal{L}_{CE-Target},
  \end{aligned}
\end{equation}
where $\mathcal{L}_{CE-Source}$ and $\mathcal{L}_{CE-Target}$ are the cross-entropy losses for the source and target domain classifiers, respectively. 

\section{Experiments}
\subsection{Dataset}
We implemented our transductive transfer learning approach using the publicly available DSIAC dataset \cite{dsiac}. This dataset is collected by the US Army Night Vision and Electronic Sensors Directorate (NVESD). The DSIAC dataset contains ten vehicle targets in the visible and MWIR domain. Among these classes, two of them are civilian vehicles (‘Pickup’, ‘Sport vehicle’), one is artillery piece (‘D20’), and the rest of the seven are military vehicles (‘2S3’, ‘BTR70’, ‘BRDM2’, ‘BMP2’, ‘MT-LB’, ‘T72’, ‘ZSU23’). The distance between target vehicles and cameras is altered from 1 kilometer to 5 kilometers. The interval of capturing distance is 500 meters. The dataset contains 189 video sequences in the MWIR domain and 97 video sequences in the visible domain. Each video sequence contains 1800 video frames. Typically, the image size in the DSIAC dataset is 640x480 pixels. 
In this experiment, we detect and crop the ATR vehicles from the DSIAC dataset by using the information from the Meta-UDA \cite{meta_domenic}. The different distances of target vehicles are projected into the canonical distance (2 kilometers) using bi-cubic interpolation. The final target chip size is 68x68x3. 

\subsection{Training Details}
Our transductive network consists of a CycleGAN and two CNN-based vehicle classifiers. The learning rate of the CycleGAN network is set to 0.0002 for the first 50 epochs; after that, the learning rate is reduced to 0.0001 for the rest of the 50 epochs. For fast convergence of the network, we have initialized the weights of the CycleGAN generator weight from the weights of summer to winter translation on the Yosemite dataset \cite{summer_winter}. The hyper-parameters of the equations \ref{eq:3}, \ref{eq:4}, \ref{eq:5} are set to $\eta_{1}$ = $\eta_{2}$ = 10, $\eta_{3}$ = $\eta_{4}$ = 5, and $\lambda_{a}$ = $\lambda_{b}$ = $\lambda_{c}$ = 1. For the initial 20 epochs of training, we set $\lambda_{CE}$ = 0.5, after that it becomes 2.5 in equation \ref{eq:7}. To optimize the weights of the generators and discriminators, we used the Adam optimizer\cite{adam} with $\beta$1 = 0.5 and $\beta$2 = 0.999. To stabilize the GAN training, we update the discriminator loss five times less than the generator loss. For training the classifiers, we also used the Adam optimizer with a learning rate of 0.0005 and $\beta$1 = 0.9, $\beta$2 = 0.999 for 40 epochs. We fine-tune the target domain classifier for 10 epochs to get the result from 1\%, 5\%, and 10\% labeled data in \autoref{fig:semi_supervised_confusion}. This experiment was conducted on an NVIDIA RTX-8000 GPU using PyTorch framework \cite{pytorch} with a batch size of 160.

\section{Results and Discussion}
\begin{figure} [ht]
   \begin{center}
   \begin{tabular}{c} %% tabular useful for creating an array of images 
   \includegraphics[scale = 0.37]{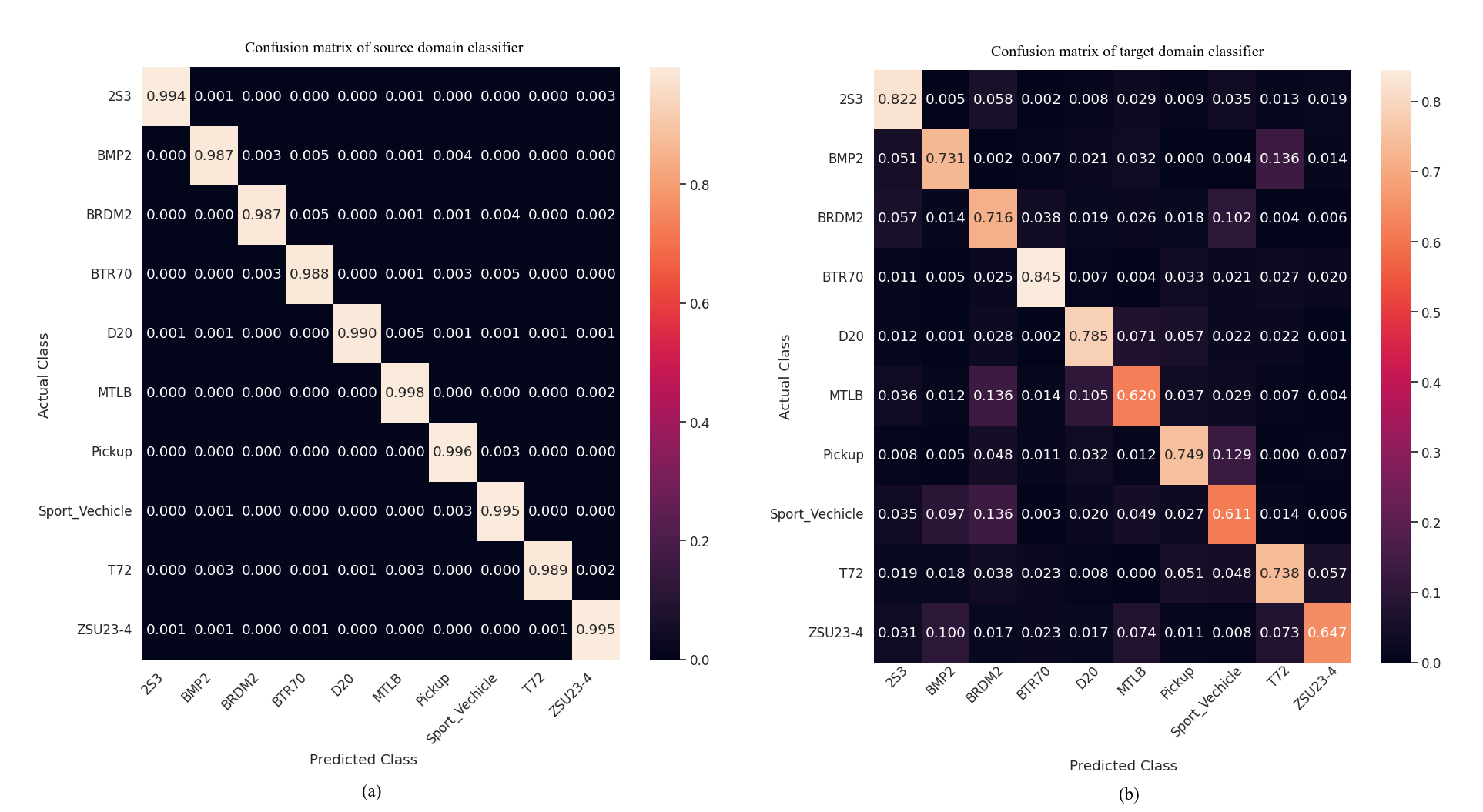}
   \end{tabular}
   \end{center}
   \caption[example] 
%>>>> use \label inside caption to get Fig. number with \ref{}
   { \label{fig:example_confusion} 
Confusion matrices of (a) the source domain classifier and (b) the target domain classifier.}
   \end{figure}

\begin{figure} [ht]
\begin{center}
\begin{tabular}{c} %% tabular useful for creating an array of images 
\includegraphics[scale = 0.37]{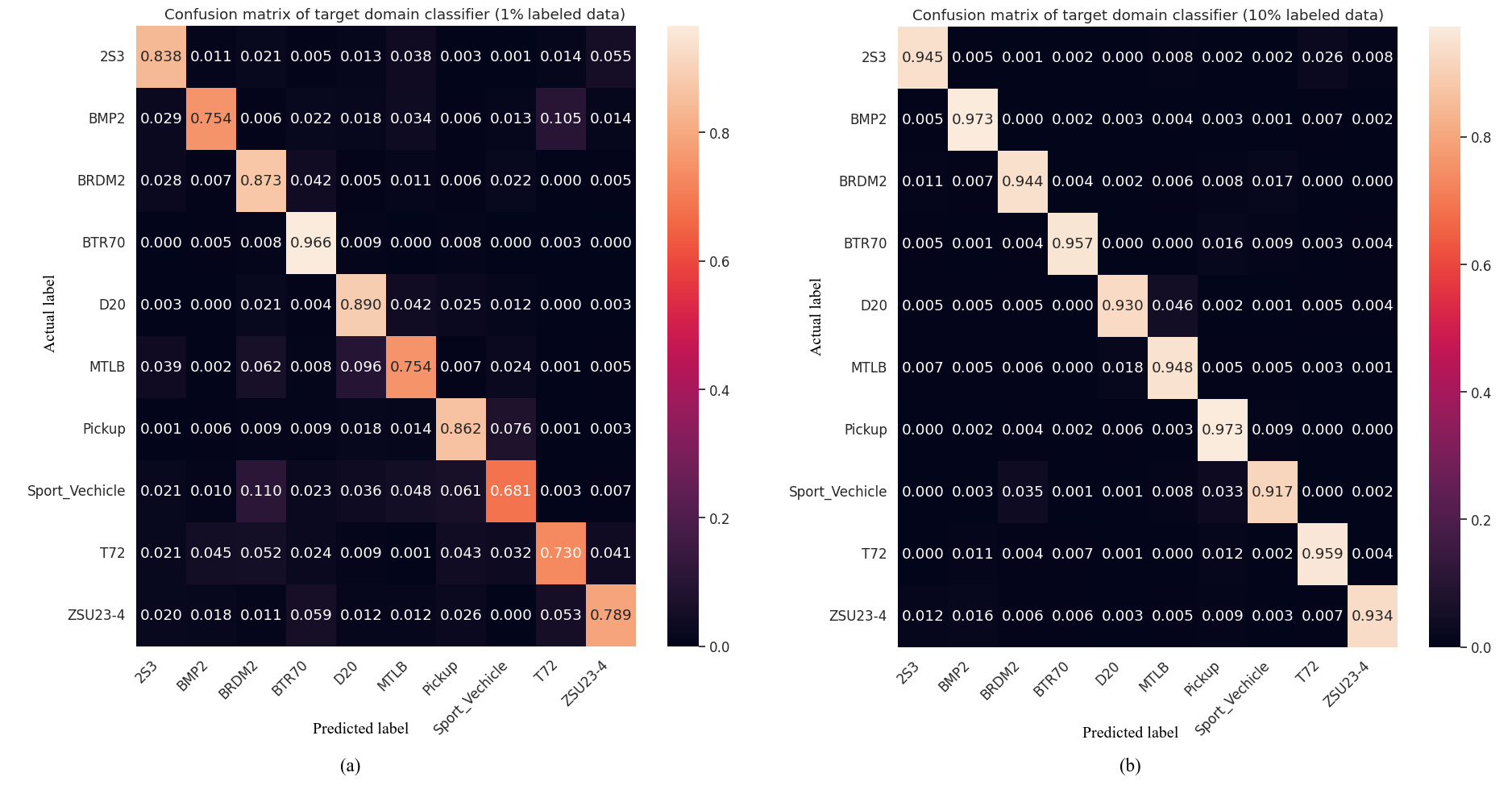}
\end{tabular}
\end{center}
\caption[example] 
%>>>> use \label inside caption to get Fig. number with \ref{}
{ \label{fig:semi_supervised_confusion} 
Confusion matrices of (a) target domain classifier (1\% labeled data) and (b) target domain classifier (10\% labeled data).}
\end{figure}

\begin{figure} [ht]
   \begin{center}
   \begin{tabular}{c} %% tabular useful for creating an array of images 
   \includegraphics[scale = 0.8]{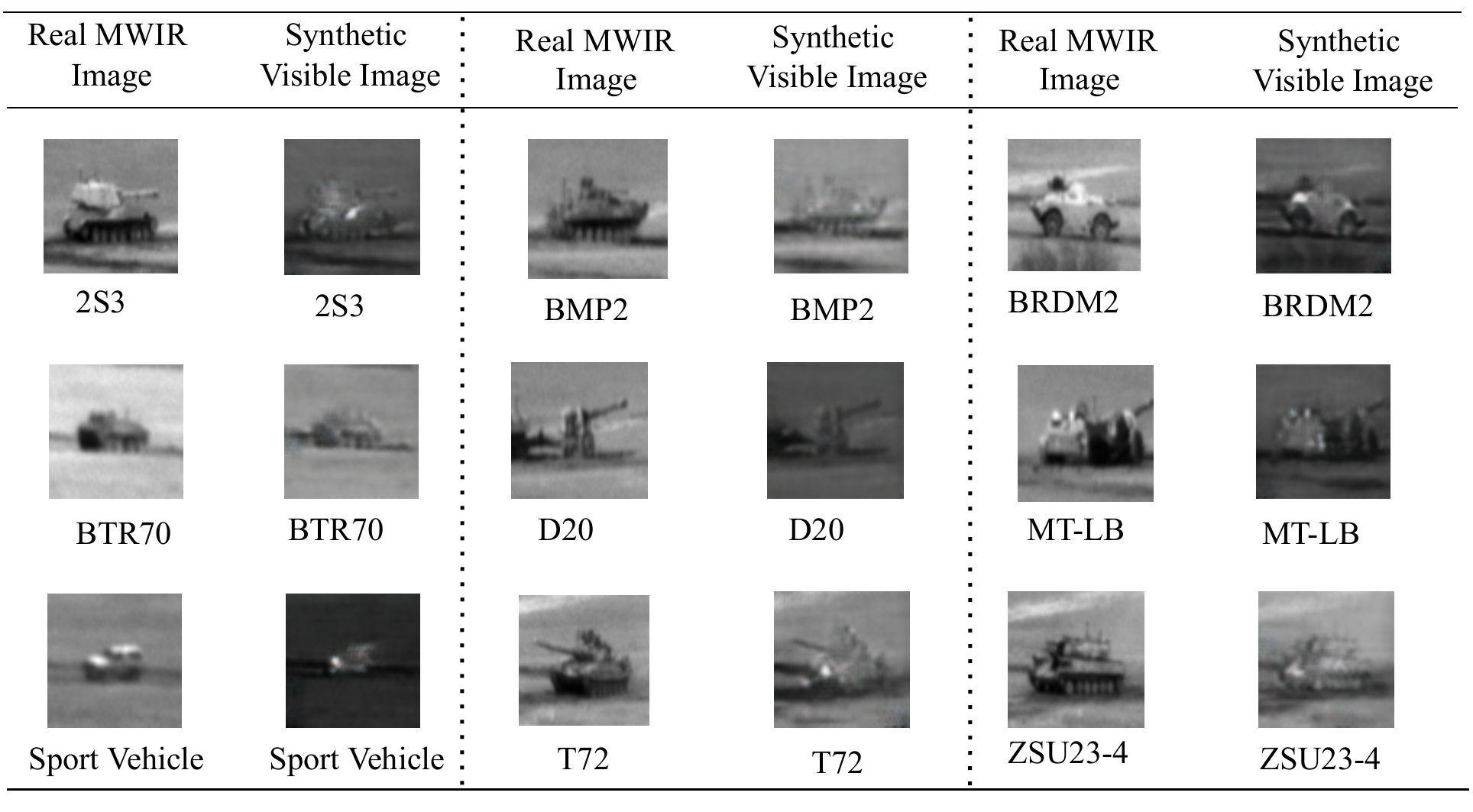}
   \end{tabular}
   \end{center}
   \caption[example] 
%>>>> use \label inside caption to get Fig. number with \ref{}
   { \label{fig:synthetic_images_transductive} 
Generated images by the proposed transductive CycleGAN framework.}
   \end{figure}

We evaluate the source and target classifier's performance on the DSIAC dataset. The training, testing, and validation datasets are constructed by randomly dividing the DSIAC dataset into 70:15:15 ratio, respectively.  The confusion matrix of the source domain classifier is illustrated in \autoref{fig:example_confusion}(a). The confusion matrix provides the normalized performance of the classifier. The average accuracy of the source domain classifier is 99.16\%. The performance of the source domain classifier with targets at different distances is investigated in \autoref{tab:distance_performance}. From this table, we can conclude that the classification performance of the source domain classifier is almost consistent when capturing images at distances between 1 kilometer to 4 kilometers. However, the source classifier performance is degraded when the target capturing distance is beyond 4 kilometers. Because long-distance target chips are low in quality as compared to short-distance ones.
We constructed a target domain classifier in this experiment using the transductive CycleGAN network. The confusion matrix of the target domain classifier is depicted in \autoref{fig:example_confusion}(b). The average accuracy of the target domain classifier is 71.6\% for the visible domain test set of the DSIAC dataset. The confusion matrix of the target domain classifier exposes that the classification performances of the ‘MTLB’, ‘Sport Vehicle’, and ‘ZSU23-4’  are lower than other classes. 
Moreover, we investigated the target domain classifier performance with a fraction of the labeled data. The accuracy of the target domain classifier is 80.24\%, 90.79\%, and 94.86\%, with 1\%, 5\%, and 10\% of the labeled target domain dataset, respectively. The confusion matrices of the visible domain classifier with 1\% and 10\% labeled data are depicted in \autoref{fig:semi_supervised_confusion}.
The synthetic images generated by transductive CycleGAN are also illustrated in \autoref{fig:synthetic_images_transductive}. The Fr\'echet Inception Distance (FID) \cite{fid} score of synthetic images is 216.109, which elicits that the network is successful in generating high-quality synthetic visible images.

Finally, the performance of the trained target domain classifier is moderate. It may be the cause of transductive CycleGAN considers bijection mappings that are too strict in condition \cite{cut}. Therefore, other unpaired domain translation methods without using bijection mappings \cite{flsesim,cut} can be employed to investigate the performance of optimal target domain classifier.
 
\begin{table}[ht]
\caption{Performance of source domain classifier with different target distance.} 
\label{tab:distance_performance}
\begin{center}       
\begin{tabular}{l l l } %% this creates two columns
%% |l|l| to left justify each column entry
%% |c|c| to center each column entry
%% use of \rule[]{}{} below opens up each row
\hline
Target Distance (m)&Accuracy(\%)&Number of Samples\\\hline
1000&99.47&4,885\\

1500&	99.72&	4,713\\
2000&	99.78&	4,895\\
2500&	99.54&	4,830\\
3000&	99.45&	4,558\\
3500&	99.76&	4,193\\
4000&	99.11&	4,506\\
4500&	97.83&	4,196\\
5000&	93.55&	4,106\\\hline

\end{tabular}
\end{center}
\end{table} 
 
\section{Conclusion}
In this paper, we propose a transductive transfer learning network consisting of a CycleGAN and two CNN-based ATR classifiers (i.e., MWIR and VIS ATR classifiers). The MWIR classifier is used to annotate the visible domain images in the DSIAC ATR dataset. Our proposed Transductive CycleGAN framework can be used to design an ATR  classifier in the target domain without any labeled data. It can also ameliorate the cumbersome and expensive manual ATR annotation process. We analyze the performance of the source domain classifier as well as the target domain classifier trained with no label data. The performance of the trained target domain classifier is 71.6\% which indicates the effectiveness of our proposed method with no labeled data in the target domain.

\acknowledgments % equivalent to \section*{ACKNOWLEDGMENTS}       
 This material is based upon work supported in part by the U. S. Army Research Laboratory and the U. S. Army Research Office under contract number: W911NF2210117.

% References
\bibliography{main} % bibliography data in report.bib
\bibliographystyle{spiebib} % makes bibtex use spiebib.bst

\end{document}